\documentclass[twocolumn]{cinc}
\usepackage{graphicx}
\usepackage{xcolor}
\usepackage{multirow} 
\usepackage{booktabs}
\usepackage{amssymb}             

\usepackage[hidelinks]{hyperref}
\begin{document}
\bibliographystyle{cinc}
\newcommand{\cmmnt}[1]{\ignorespaces}
\graphicspath{{images/}{../images/}}


\title{Dual Bayesian ResNet: A Deep Learning Approach to Heart Murmur Detection}


\author {Benjamin Walker$^{1}$, Felix Krones$^{2}$, Ivan Kiskin$^{3, 4}$, Guy Parsons$^{5}$, Terry Lyons$^{1}$,  Adam Mahdi$^{2,3}$ \\
\ \\ 
$^1$ Mathematical Institute, University of Oxford, Oxford, UK,
$^2$ Oxford Internet Institute, University of Oxford, Oxford, UK, 
$^3$  People-Centred AI Institute, University of Surrey, Surrey, UK, $^4$ Surrey Sleep Research Centre, University of Surrey, Surrey, UK, $^5$  Intensive Care Registrar, Thames Valley Deanery, NIHR Academic Clinical Fellow at University of Oxford, Oxford, UK.}

\maketitle

\begin{abstract}
    This study presents our team PathToMyHeart's contribution to the George B. Moody PhysioNet Challenge 2022.
    Two models are implemented. The first model is a Dual Bayesian ResNet (DBRes), where each patient's recording is segmented into overlapping log mel spectrograms. These undergo two binary classifications: present versus unknown or absent, and unknown versus present or absent. The classifications are aggregated to give a patient's final classification. The second model is the output of DBRes integrated with demographic data and signal features using XGBoost.
    DBRes achieved our best weighted accuracy of $0.771$ on the hidden test set for murmur classification, which placed us fourth for the murmur task. (On the clinical outcome task, which we neglected, we scored 17th with costs of 12637.) On our held-out subset of the training set, integrating the demographic data and signal features improved DBRes's accuracy from $0.762$ to $0.820$. However, this decreased DBRes's weighted accuracy from $0.780$ to $0.749$. 
    Our results demonstrate that log mel spectrograms are an effective representation of heart sound recordings, Bayesian networks provide strong supervised classification performance, and treating the ternary classification as two binary classifications increases performance on the weighted accuracy.


\end{abstract}

\section{Introduction} 
Congenital heart disease occurs when there are problems with the early development of the heart's structure. The detection of heart murmurs (sounds made by turbulent blood flow through the heart) and other noises with this technique can indicate structural defects in the heart. The analysis of signals in early life may therefore provide a rapid and non-invasive screening test for the presence of cardiac structural defects, enabling early diagnosis and intervention \cite{frank2011evaluation}. 



In this work, we present our contribution to the heart murmur classification task from the 2022 George B. Moody PhysioNet Challenge \cite{goldberger2000physiobank}. The aim was to design an open-source algorithm to classify the presence, absence, or unknown cases of heart murmurs from heart sound recordings\footnote{Our algorithm is publicly available at https://github.com/Benjamin-Walker/heart-murmur-detection}. Our contributions are threefold: (i) we design and implement two deep learning modelling approaches, one trained purely on two-dimensional representations of data (spectrograms) derived from the heart sound recordings, and a second that additionally utilises a patient's demographic data and features derived directly from the sound recordings; (ii) we compare the relative contribution of different data modalities to heart murmur classification; and 
(iii) discuss potential mechanisms and implications for the observed results.

\section{Methodology}

\subsection{Data}
\begin{figure}[htbp]
\includegraphics[width=7.5cm]{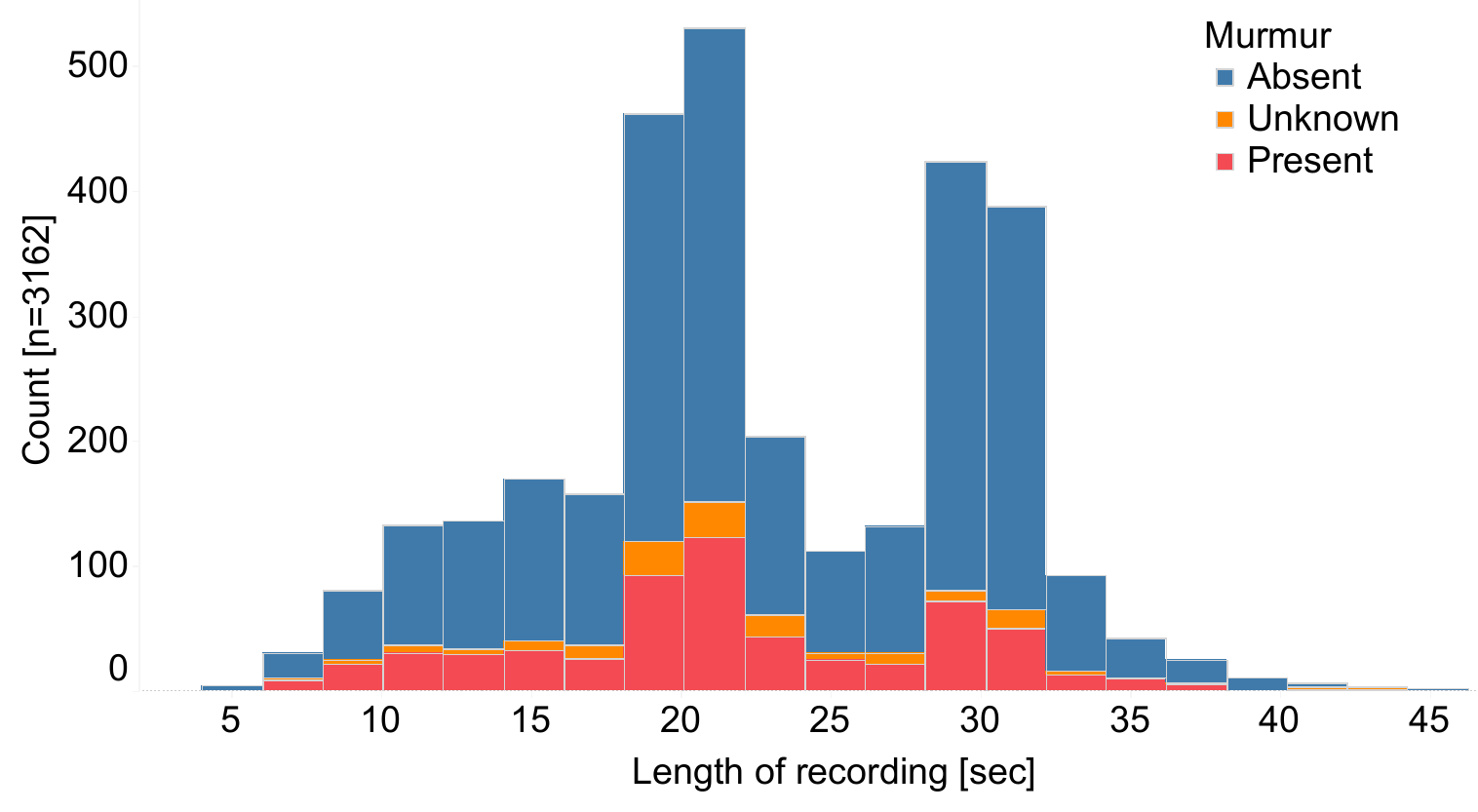}
\caption{Histogram of recording lengths with the corresponding murmur label.}
\label{fig:rec_length}
\end{figure}

The data provided for this year's challenge was collected from two screening campaigns in Northeast Brazil in July/August 2014 and June/July 2015 \cite{oliveira2021circor}. The dataset contains heart sound recordings of length $5$ to $45$ seconds (Figure~\ref{fig:rec_length}), together with demographic data, which consists of age categories, sex, height, weight, and pregnancy status. There are $1568$ patients in the dataset, of which $60\,\%$ $(942)$ were given to the participants for training. For each patient, there were up to six heart sound recordings available, with $5272$ recordings in the full dataset and $3163$ in the training set. Each recording was taken from one of these locations: pulmonary valve, aortic valve, mitral valve, tricuspid valve, or other. Each patient has a heart murmur label, which can be present, unknown, or absent.


\subsection{Scores} 
The murmur challenge score used to evaluate classifiers was the weighted accuracy 
\begin{equation}\label{eq:accu}
    s = \frac{5c_p + 3c_u + c_a}{5t_p + 3t_u + t_a},
\end{equation}
where $c_i$ is the number of correct classifications and $t_i$ is the total number of cases for present ($i=p$), unknown ($i=u$), and absent ($i=a$). Models were also compared locally using the accuracy $({c_p + c_u + c_a})/({t_p + t_u + t_a})$ and the per class accuracies ${c_i}/{t_i}$.



\subsection{Data preparation} 

\begin{figure}[htbp]
\includegraphics[width=7.5cm]{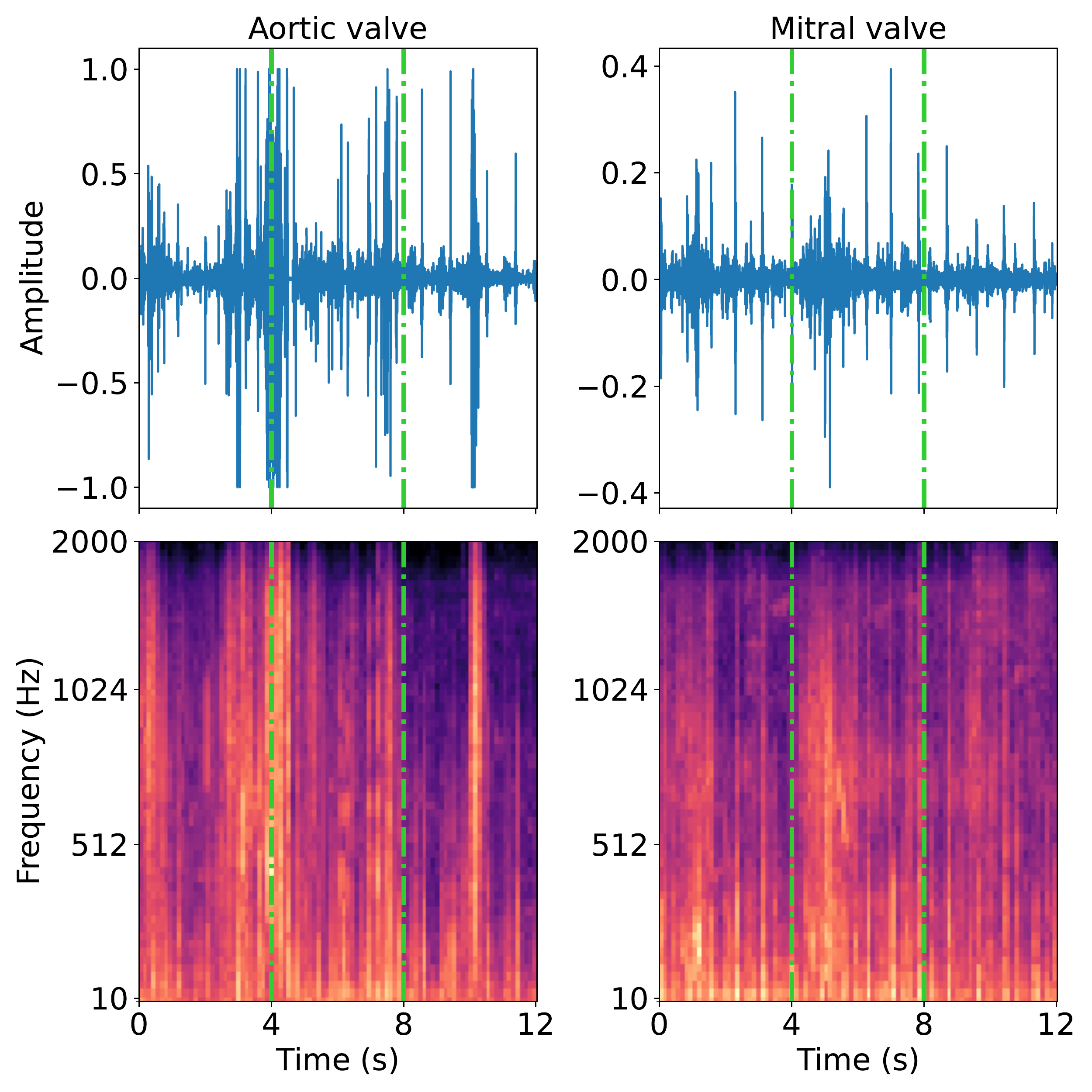}
\caption{Example heart sound recordings (top row) for a patient 
with \textit{present} murmur recorded at the aortic valve (left column) and mitral valve (right column). The bottom row shows the log mel spectrogram, as parameterised in the code. The dash-dotted lines show how the data was partitioned into $4$ second two-dimensional inputs.}
\label{fig:mel_spec}
\end{figure}

The short-time Fourier transform determines the frequency and phase component of sections of a signal as it varies over time. This is achieved using a windowed Fourier transform \cite{SEJDIC2009153}. Its result is often represented as a spectrogram. Log mel spectrograms further map the frequency axis to the (logarithmic) mel scale, which aims to maintain the distance humans perceive between pitches (cf. Figure~\ref{fig:mel_spec}). This representation has repeatedly shown success in recent audio classification challenges tasks, either as standalone features, or used in combination with other acoustic features \cite{Challenge,dcasetask42021}.

The audio data was prepared by computing log mel spectrograms and extracting signal features from the recordings. The signal features extracted include summary features in the time and frequency domains, as well as summary features of the spectral centroid, rolloff and bandwidth. These features were extracted using the Python packages \textsc{scipy} and \textsc{librosa}. 

When calculating the spectrograms, each recording is divided into overlapping segments using a window of length $4$ seconds and stride of $1$ second. For each segment, individual spectrograms with $64$ coefficients are calculated using a Fast Fourier Transform with a periodic Hanning window of length $25$ milliseconds and stride $10$ milliseconds. The spectrogram's minimum and maximum frequencies were $10$ and $2000\,\rm{Hz}$, respectively.

The demographic data was preprocessed using the example code provided by organisers of this year's Challenge. This consisted of converting age labels into approximate ages in months, one hot encoding the sex label, and converting the pregnancy status into a binary variable. Missing values are handled using a mean imputation.

\begin{figure}[htbp]
\includegraphics[width=7.8cm]{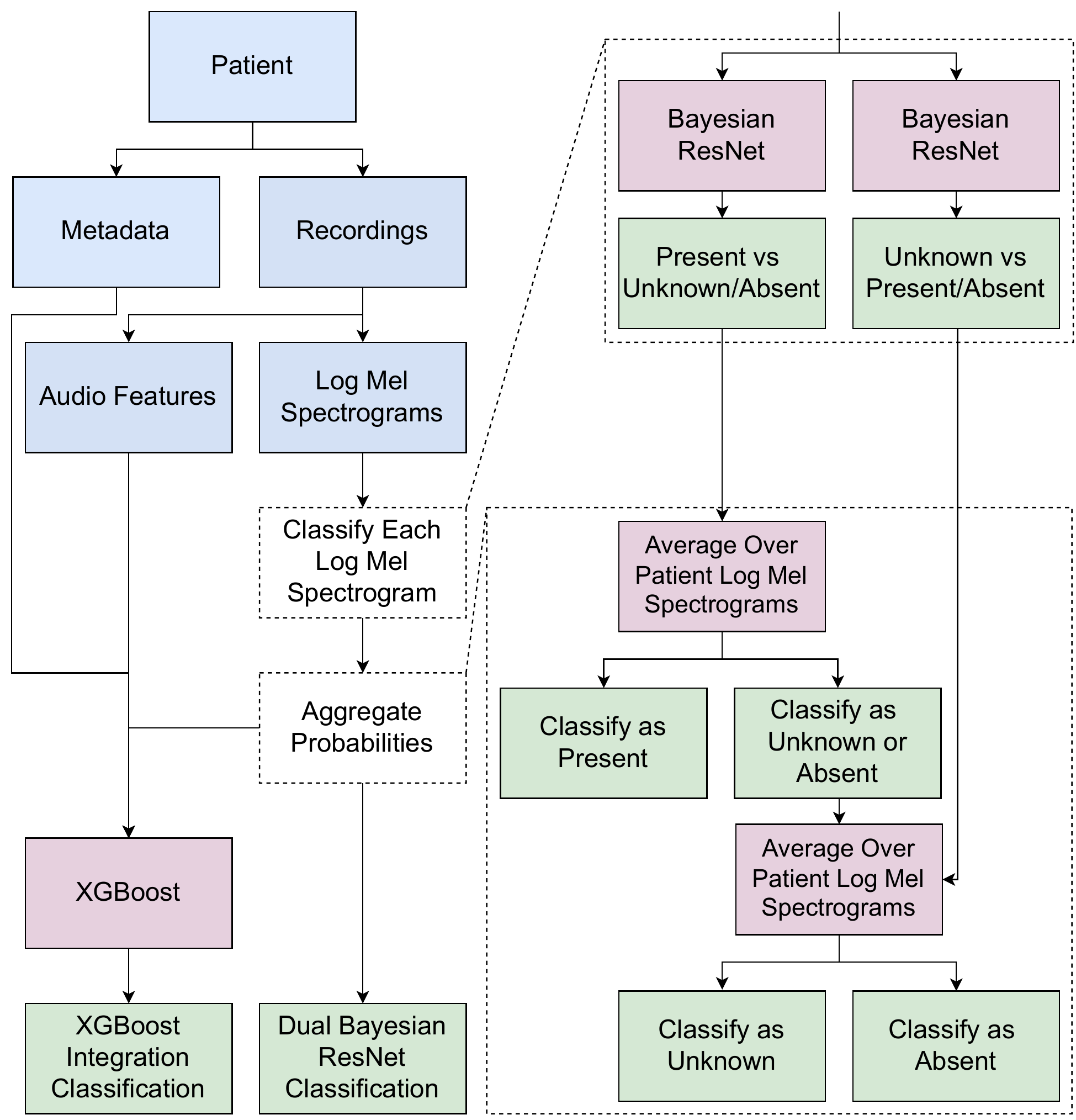}
\caption{A schematic diagram of the Dual Bayesian ResNet (DBRes) and XGBoost integration models. The blue, purple, and green boxes represent data, models and classifications, respectively.}
\label{fig:architecture}
\end{figure}


\subsection{Models} 
{\bf Bayesian Neural Networks.} 
Bayesian neural networks provide strong performance for supervised classification tasks and an estimate of the uncertainty in a classification, both of which are desirable for real-data tasks such as this challenge. Additionally, they naturally suit Bayesian decision theory. This benefits decision-making applications where the associated cost is dependent on the type of incorrect decision, such as the weighted accuracy given in (\ref{eq:accu}) \cite{cobb2018loss}. The core of the audio-based inference is performed using two
Monte Carlo dropout ResNet50 Bayesian neural networks. ResNet has achieved state-of-the-art performance in audio tasks \cite{palanisamy2020rethinking} motivating its use as a dependable baseline model in our study. In order to approximate the model posterior as test time, dropout layers are added to the modules \texttt{BasicBlock()} and \texttt{Bottleneck()}, as well as the overall model construction. The model was pre-trained on ImageNet and the layers were re-trainable. For more details see \cite[Appendix B.4]{kiskin2021humbugdb}.


{\bf Dual Bayesian ResNet and XGBoost integration.} Figure~\ref{fig:architecture} is a schematic diagram of the data preparation and two models considered in this paper, the Dual Bayesian ResNet (DBRes) and the DBRes with demographic data and signal feature integration via XGBoost. There are three major components to the models: classifying the individual spectrograms, aggregating these classifications, and integrating the demographic data and signal features via XGBoost \cite{pimentel2019uncertainty}.

The ternary murmur classification is split into two binary classifications: present versus unknown or absent, and unknown versus present or absent. Separate Bayesian ResNet50 networks are trained on the individual spectrograms for each of these tasks. During testing, a patient's individual spectrograms are simultaneously classified using both networks.

The individual spectrograms classifications are aggregated by first taking the arithmetic mean of the output from the present versus unknown or absent ResNet50. If this averaged output classifies the patient's murmur as present, it is classified as present. If not, then the arithmetic mean of the output from the unknown versus present or absent ResNet50 is taken. If this averaged output classifies the patient's murmur as unknown, it is classified as unknown, else the patient's murmur is classified as absent. DBRes is the outcome from this classification aggregation.

This choice of model structure and classification aggregation was chosen to prioritise the accuracy of present classifications, aligning with the priority in the murmur challenge score (\ref{eq:accu}). 

The second model considered in this paper integrates the output from the DBRes with the patient's demographic data and extracted signal features using XGBoost. This model is referred to as DBRes with XGBoost integration.

\begin{table}[htbp]
\caption{Murmur labels by age [n (\% of 942)].}
\smallskip
\label{tab:murmur_outcome}
\small\renewcommand{\arraystretch}{1.2}
\begin{tabular}{llllr} \hline
            &Absent        &Unknown       &Present       &Sum \\
            \hline\hline
Neonate     & 4 (0.4)       & 1 (0.1)       & 1 (0.1)       & 6 (0.6) \\
Infant      & 76 (8.1)      & 25 (2.7)      & 25 (2.7)      & 126 (13.4) \\
Child       & 495 (52.6)    & 37 (3.9)      & 132 (14.0)    & 664 (70.5) \\
Adolescent  & 53 (5.6)      & 3 (0.3)       & 16 (1.7)      & 72 (7.6) \\
Missing     & 67 (7.1)      & 2 (0.2)       & 5 (0.5)       & 74 (7.9) \\
\hline\hline
Sum        & 695 (73.8)    & 68 (7.2)      & 179 (19.0)    & 942 (100) \\\hline
\bottomrule
\end{tabular}
\end{table}

\subsection{Code availability}
Our code is available on a GitHub repository \cite{Walker22}.

\section{Results}
{\bf Preliminary data analysis.} The data contain mainly children and is highly unbalanced, with murmurs being absent in $74\%$ of patients, present in $19\%$ of patients, and unknown in $7\%$ of patients, as shown in Table~\ref{tab:murmur_outcome}. 


{\bf Models performance.} Table~\ref{tab:scores} provides the per class accuracy, accuracy, and the murmur challenge score for DBRes and DBRes with XGBoost integration when evaluated on our held-out subset of the training set. DBRes scored $0.771$ (4th place) when evaluating the murmur challenge score using the PhysioNet hidden test set (cf. Table \ref{tab:murmur-scores}). On the clinical outcome task, which we neglected, we scored 17th with costs of 12637 (cf. Table \ref{tab:outcome-scores}). The similarity between the murmur challenge score on our held-out subset of the training set and the PhysioNet hidden test set demonstrate that the held-out subset of the training set has been constructed in a sound way to promote model generalisation across datasets.


\begin{table*}
\caption{Murmur metric scores on held-out subset of the training set for DBRes and DBRes with XGBoost Integration.}\label{tab:scores}
\smallskip
\centering\small\renewcommand{\arraystretch}{1.2}
\begin{tabular}{lccccccc} 
\hline
& \multirow{2}{*}{Features} & \multicolumn{3}{c}{Accuracy per Class}& \multirow{2}{*}{Accuracy} & \multirow{2}*{\parbox{25mm}{\centering Murmur \\ Challenge Score}} \\ 
&& Present                               & Unknown                    & Absent  & & \\ \hline\hline
DBRes & Spectrograms & $0.833$ & $0.615$ & $0.757$ & $0.762$ & $0.780$ \\
DBRes with XGBoost & Spectrograms, signal features, demographic data  & $0.750$ & $0.231$ & $0.893$ & $0.820$ & $0.749$ \\ 
\hline
\end{tabular}
\end{table*}

\begin{table}[tbp]
    \centering
    \begin{tabular}{r|r|r|r}
        Training        & Validation & Test & Ranking \\\hline
        0.78 &       0.768 & 0.771 &  4/40 \\\hline
    \end{tabular}
    \caption{Weighted accuracy metric scores for our final selected entry for the murmur detection task. Training score is on a held-out subset of the public training set, repeated scoring is used on the hidden validation set, and one-time scoring is used on the hidden test set.}
    \label{tab:murmur-scores}
\end{table}

\begin{table}[tbp]
    \centering
    \begin{tabular}{r|r|r|r}
        Training        & Validation & Test  & Ranking \\\hline
        13023 &      10411 & 12637 &  17/39 \\\hline
    \end{tabular}
    \caption{Cost metric scores for our final selected entry for the clinical outcome identification task. Training score is on a held-out subset of the public training set, repeated scoring is used on the hidden validation set, and one-time scoring is used on the hidden test set.}
    \label{tab:outcome-scores}
\end{table}

\section{Discussion}
{\bf Principal findings.} 
In this work we proposed and evaluated two Bayesian deep learning approaches to classifying murmurs as present, absent, or unknown from heart sound recordings and demographic data. The first approach, DBRes, implements two binary Bayesian ResNet50 networks, which classify murmurs in segmented spectrograms of heart sound recordings. 
The second approach combines the output from DBRes with features extracted from audio signals and patients' demographic data via XGBoost. 


The results in Table~\ref{tab:scores} show that spectrograms are a good representation of the data, and  when combined with ResNet provide the majority of the predictive power. Furthermore, the integration of demographic data and signal features improves the accuracy. However, this integration decreases the weighted accuracy.

{\bf Potential mechanisms and implications.} 
The results demonstrate that the architecture of DBRes prioritises the accuracy on present, then on unknown, and then on absent cases, without using a weighted loss function. This is further demonstrated in Table~\ref{tab:scores}, as present has the highest per class accuracy for DBRes. In contrast, the current implementation of XGBoost integration is not optimised for the weighted accuracy. Therefore, the accuracy of absent cases is prioritised, as they comprise the majority of the dataset. This leads to the observed decrease in the weighted accuracy.


Our results demonstrate the viability of a deep neural network approach to classifying heart murmurs from heart sound recordings. Given further work to improve the specificity of murmur classification, these models could be a component of an algorithmic screening for congenital heart disease.

{\bf Future research.} 
A direction of future research is developing a superior method for integrating patients' demographic data and signal features with the outputs from DBRes. This could include adjusting the objective function of XGBoost to align with the challenge score or investigating other methods of multimodal data fusion  \cite{duvieusart2022multimodal}. 



\section*{Acknowledgement}
Benjamin Walker was funded by the Hong Kong Innovation and Technology Commission (InnoHK Project CIMDA). Terry Lyons was funded in part by the EPSRC [grant number EP/S026347/1], in part by The Alan Turing Institute under the EPSRC grant EP/N510129/1, the Data Centric Engineering Programme (under the Lloyd’s Register Foundation grant G0095), the Defence and Security Programme (funded by the UK Government) and the Office for National Statistics \& The Alan Turing Institute (strategic partnership) and in part by the Hong Kong Innovation and Technology Commission (InnoHK Project CIMDA).

%
\bibliography{refs}

\begin{thebibliography}{10}
\expandafter\ifx\csname url\endcsname\relax
  \def\url#1{\texttt{#1}}\fi
\expandafter\ifx\csname urlprefix\endcsname\relax\def\urlprefix{URL }\fi

\bibitem{frank2011evaluation}
Frank JE, Jacobe KM.
\newblock Evaluation and management of heart murmurs in children.
\newblock American family physician 2011;\hspace{0pt}84(7):793--800.

\bibitem{goldberger2000physiobank}
Goldberger AL, Amaral LA, Glass L, Hausdorff JM, Ivanov PC, Mark RG, et~al.
\newblock Physiobank, physiotoolkit, and physionet: components of a new
  research resource for complex physiologic signals.
\newblock circulation 2000;\hspace{0pt}101(23):e215--e220.

\bibitem{oliveira2021circor}
Oliveira J, Renna F, et~al.
\newblock The circor digiscope dataset: from murmur detection to murmur
  classification.
\newblock IEEE Journal of Biomedical and Health Informatics
  2021;\hspace{0pt}26(6):2524--2535.

\bibitem{SEJDIC2009153}
Sejdić E, Djurović E, Jiang J.
\newblock Time–frequency feature representation using energy concentration:
  An overview of recent advances.
\newblock Digital Signal Processing 2009;\hspace{0pt}19:153--183.

\bibitem{Challenge}
Reyna MA, Kiarashi Y, Elola A, Oliveira J, Renna F, Gu A, et~al.
\newblock Heart murmur detection from phonocardiogram recordings: The {G}eorge
  {B}. {M}oody {P}hysionet {C}hallenge 2022.

\bibitem{dcasetask42021}
Wisdom S, Erdogan H, et~al.
\newblock {DCASE 2021 Task 4: Sound event detection and separation in domestic
  environments}, 2021.

\bibitem{cobb2018loss}
Cobb AD, Roberts SJ, Gal Y.
\newblock Loss-calibrated approximate inference in {B}ayesian neural networks.
\newblock arXiv180503901 2018;\hspace{0pt}.

\bibitem{palanisamy2020rethinking}
Palanisamy K, Singhania D, Yao A.
\newblock {Rethinking CNN models for audio classification}.
\newblock arXiv200711154 2020;\hspace{0pt}.

\bibitem{kiskin2021humbugdb}
Kiskin I, Sinka M, et~al.
\newblock {HumBugDB: A Large-scale Acoustic Mosquito Dataset}.
\newblock In Thirty-fifth Conference on Neural Information Processing Systems
  Datasets and Benchmarks Track (Round 2). 2021;\hspace{0pt} --.

\bibitem{pimentel2019uncertainty}
Pimentel MAF, Mahdi A, Redfern O, Santos MD, Tarassenko L.
\newblock Uncertainty-aware model for reliable prediction of sepsis in the
  {ICU}.
\newblock In 2019 Computing in Cardiology (CinC). 2019;\hspace{0pt} 1--4.

\bibitem{Walker22}
Walker B, Krones F, Kiskin I, Parsons G, Lyons T, Mahdi A.
\newblock Dual {B}ayesian {R}es{N}et: A {P}ython code for heart murmur
  detection. {G}it{H}ub repository.
\newblock \url{https://github.com/Benjamin-Walker/PhysionetChallenge2022},
  2022.

\bibitem{duvieusart2022multimodal}
Duvieusart B, Krones F, Parsons G, Tarassenko L, Papie{\.z} B, Mahdi A.
\newblock Multimodal cardiomegaly classification with image-derived digital
  biomarkers.
\newblock In Medical Image Understanding and Analysis. 2022;\hspace{0pt}
  13--27.

\end{thebibliography}
\medskip
{Address for correspondence:}
\smallskip

\begin{minipage}{\textwidth}
{\small
\begin{tabular}{lll}
Benjamin Walker                     &\qquad &  Adam Mahdi\\
35 Hart Street                      &&  1 St Giles\\ 
Oxford, OX26BN, UK                  && Oxford, OX23JS, UK \\
 benjamin.walker1@maths.ox.ac.uk    && adam.mahdi@oii.ox.ac.uk 
\end{tabular}
}
\end{minipage}


\end{document}